\documentclass[conference]{IEEEtran}
\IEEEoverridecommandlockouts
\usepackage{cite}
\usepackage{amsmath,amssymb,amsfonts}
\usepackage{algorithmic}
\usepackage{graphicx}
\usepackage{textcomp}
\usepackage{xcolor}
\usepackage{multirow}
\usepackage{siunitx} % 引入siunitx包
\usepackage{subcaption}
\usepackage{float}
\usepackage{authblk}
\usepackage{booktabs}

\def\BibTeX{{\rm B\kern-.05em{\sc i\kern-.025em b}\kern-.08em
    T\kern-.1667em\lower.7ex\hbox{E}\kern-.125emX}}
\begin{document}

% \title{TSFDT-SSVEP: A Temporal-Spectrum Fusion Dual-Stream Transformer for SSVEP Decoding}

% ZS
\title{
% SSVEP-BiMA: Bifocal Masking Attention Leveraging Symmetric-Antisymmetric Components and Native Time Series for Robust SSVEP Decoding
SSVEP-BiMA: Bifocal Masking Attention Leveraging Native and Symmetric-Antisymmetric Components for Robust SSVEP Decoding

}

\author{
Yuxin Liu\textsuperscript{1}, Zhenxi Song\textsuperscript{1}$^{*}$, Guoyang Xu\textsuperscript{1}, Zirui Wang\textsuperscript{1}, Feng Wan\textsuperscript{2},Yong Hu\textsuperscript{2}, Min Zhang\textsuperscript{1}, and Zhiguo Zhang\textsuperscript{1} \\
\textsuperscript{1}Harbin Institute of Technology, Shenzhen, China \\
\textsuperscript{2}University of Macau, Macau, China \\
\textsuperscript{3}The University of Hong Kong, Hong Kong, China \\
% \thanks{\textsuperscript{†}Co-first authors.}
\vspace{-1.5em} 
\thanks{ 
 \noindent\rule{0.5\linewidth}{0.25pt} 
 \\ 
$^{*}$Corresponding author: Zhenxi Song (songzhenxi@hit.edu.cn). This work is supported by the National Natural Science Foundation of China (Grant No. 62306089), the Shenzhen Science and Technology Program (Grant No. RCBS20231211090800003).}
} 

\maketitle

\begin{abstract}
Brain-computer interface (BCI) based on steady-state visual evoked potentials (SSVEP) is a popular paradigm for its simplicity and high information transfer rate (ITR). Accurate and fast SSVEP decoding is crucial for reliable BCI performance. However, conventional decoding methods demand longer time windows, and deep learning models typically require subject-specific fine-tuning, leaving challenges in achieving optimal performance in cross-subject settings. This paper proposed a biofocal masking attention-based method (SSVEP-BiMA) that synergistically leverages the native and symmetric-antisymmetric components for decoding SSVEP. By utilizing multiple signal representations, the network is able to integrate features from a wider range of sample perspectives, leading to more generalized and comprehensive feature learning, which enhances both prediction accuracy and robustness. We performed experiments on two public datasets, and the results demonstrate that our proposed method surpasses baseline approaches in both accuracy and ITR. We believe that this work will contribute to the development of more efficient SSVEP-based BCI systems.
\end{abstract}
\begin{IEEEkeywords}
Brain-computer Interface, Steady-state Visual Evoked Potential, Transformer, Dual-view Strategy, Masking Self-attention.
\end{IEEEkeywords}

\section{Introduction}
Brain-computer interface (BCI) has become a popular research area \cite{b1}, offering new possibilities for human-computer interaction. Steady-state visual evoked potential (SSVEP), which are electroencephalographic activity generated in the brain when an individual focuses on a rapidly flickering light source or other visual stimuli, are widely used to encode user intentions due to its ease of use and high information transfer rates. The applications of SSVEP-based BCI include spellers \cite{b2},\cite{b3},\cite{b4}, gaming \cite{b5},\cite{b6} and device control \cite{b7},\cite{b8}. 

In BCI systems, decoding errors can result in costly mistakes, making the precise identification of SSVEP signals crucial. Previous studies have proposed several classical and state-of-the-art methods based on statistical analysis, machine learning, or deep learning. However, statistical methods \cite{b9}, \cite{b10}, \cite{b11}, \cite{b12}, \cite{b13} without training data generally require long time windows to be effective. Meanwhile, most learning algorithms \cite{b14}, \cite{b15}, \cite{b16}, \cite{b17}, \cite{b18} depend on subject-specific calibration for improved performance, which is a time-consuming process limiting BCI applications. 

SSVEP can be represented as time-series data or spectrum. Both temporal and spectral components of SSVEP have been shown to play important roles in SSVEP classification tasks in previous work. EEGNet \cite{b19} uses a compact CNN for SSVEP and other types of EEG signal classification. SSVEPNet \cite{b20} combines CNN and LSTM for temporal data processing with label smoothing. CCNN \cite{b21} focuses on spectral data using CNN and FFT, while SSVEPformer \cite{b22} combines CNN and MLP for spectrum-based feature encoding and attention learning. However, few studies have explored using both temporal and spectral features to enhance decoding performance. TFF-former \cite{b23} was the first to propose time-frequency fusion for SSVEP tasks but only considered amplitude, neglecting other important spectral features such as phases, limiting its performance. Its complex architecture also struggled with smaller datasets, reducing efficiency and practicality in BCI systems. It is necessary to propose a method that fully leverages diverse information, achieves high training and computational efficiency, and maintains strong accuracy and robustness in cross-subject and limited-data scenarios.

\begin{figure*}[t]
	\centering
	\includegraphics[width=\textwidth]{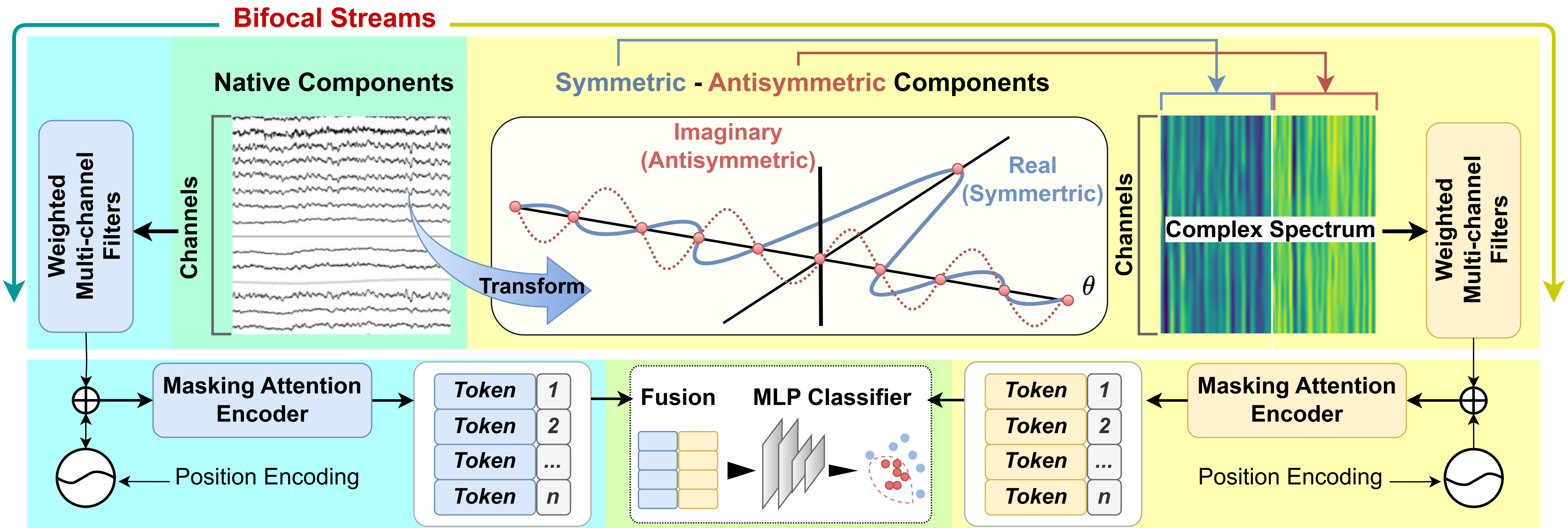} % 替换为实际图片路径
	\caption{The framework of SSVEP-BiMA.}
\end{figure*}
In this paper, we proposed a novel and compact bifocal masking attention-based model, SSVEP-BiMA, which synergically leverages the distributions of symmetric and antisymmetric components along with the original time series data for decoding SSVEP. This new approach concurrently processes and integrates these distinct signal representations, thereby enhancing the robustness and accuracy of SSVEP decoding. By incorporating both original and Fourier-transformed components encompassing amplitude and phase information, the model provides a more completed analysis framework that effectively captures the inherent dynamics of the SSVEP signals. With the compact structure, SSVEP-BiMA achieves faster convergence with less data, reducing the risk of overfitting while accelerating prediction speed and minimizing resource consumption, making it more suitable for practical BCI systems. Our contributions can be summarized as:

(1) We proposed a compact network (SSVEP-BiMA) with a bifocal attention mechanism that, for the first time, integrates native EEG and symmetric-antisymmetric components rich in amplitude and phase information for SSVEP decoding. With multiple signal representations, the method enhances neural response interpretability and decoding performance, offering deeper insights into brain activity linked to visual stimuli.
% -ZS
%（1）

(2) We further introduced a convolution-based multi-channel filter and masking attention mechanism to filter out noise and irrelevant information while improving the attention's aggregation capability. With the integration of multi-channel filter, masking attention and dual-view strategy, the model's robustness is further enhanced.

% -ZS
%（2）

(3) We conducted extensive cross-subject experiments on two public SSVEP datasets, showing that SSVEP-BiMA outperforms baseline methods and is well-suited for data-scarce, cross-subject, and low-computation SSVEP decoding scenarios, which facilitates the development of BCI systems. 

% -ZS
%（3）

\section{Method}
\subsection{Overview}
As illustrated in Fig.1, native components of EEG data are transformed into complex spectra consisting of the symmetric and antisymmetric components in the complex spectral representation block. Next, the native and symmetric-antisymmetric components are fed into the bifocal Transformer streams which consist of three modules: weighted multi-channel filter (WMF), position encoding, and masking multi-head self-attention encoder. These modules will be explained in detail in the following sections. The outputs of the dual transformer streams are fused at the feature level and classified using an multilayer perceptron (MLP) head. To achieve more efficient computation and facilitate training for real-time EEG decoding tasks, we designed the model to be as simple and compact as possible, enabling easier convergence, faster inference, and reduced resource consumption.
\subsection{Complex Spectrum Representation}
The original EEG data can be transformed into the frequency domain using the Fast Fourier Transform (FFT). Inspired by \cite{b21}, we extracted the sine and cosine components of the signal at the corresponding frequencies, representing the symmetric and antisymmetric characteristics of the signal, respectively, which can be formulated as:
\begin{equation}
	I_{\text{CSF}} = 
	\begin{bmatrix}
		\text{Real}(FFT(x_{\text{ch}1}))\|\text{Imag}(FFT(x_{\text{ch1}})) \\
		\text{Real}(FFT(x_{\text{ch}2}))\|\text{Imag}(FFT(x_{\text{ch2}})) \\
		\vdots \\
		\text{Real}(FFT(x_{\text{ch}n}))\|\text{Imag}(FFT(x_{\text{ch}n})) \\
	\end{bmatrix}
\end{equation}
where $\text{I}_\text{CSF}$ denotes the complex spectrum features, symbol $\|$ denotes the concatenate operation, and $x_{\text{ch}n}$ denotes the native EEG data of channel n. Compared to amplitude spectrum features, these components include not only amplitude information but also phase information, which has been shown in \cite{b24},\cite{b25},\cite{b26} to play a crucial role in SSVEP classification. SSVEP signals are widely recognized as phase-locked signals, meaning that for a given visual stimulation frequency, there is a fixed SSVEP response phase. In multi-electrode EEG systems, analyzing the phase relationships between different channels can uncover hidden dependencies between signals, thereby enhancing classification performance, especially in cases where noise is significant or amplitude information is insufficient. Frequency-domain features incorporating phase information have been shown to outperform amplitude-only features in the experiments conducted in \cite{b21}.
\begin{figure*}[t]
	\centering
	\includegraphics[width=0.9\textwidth]{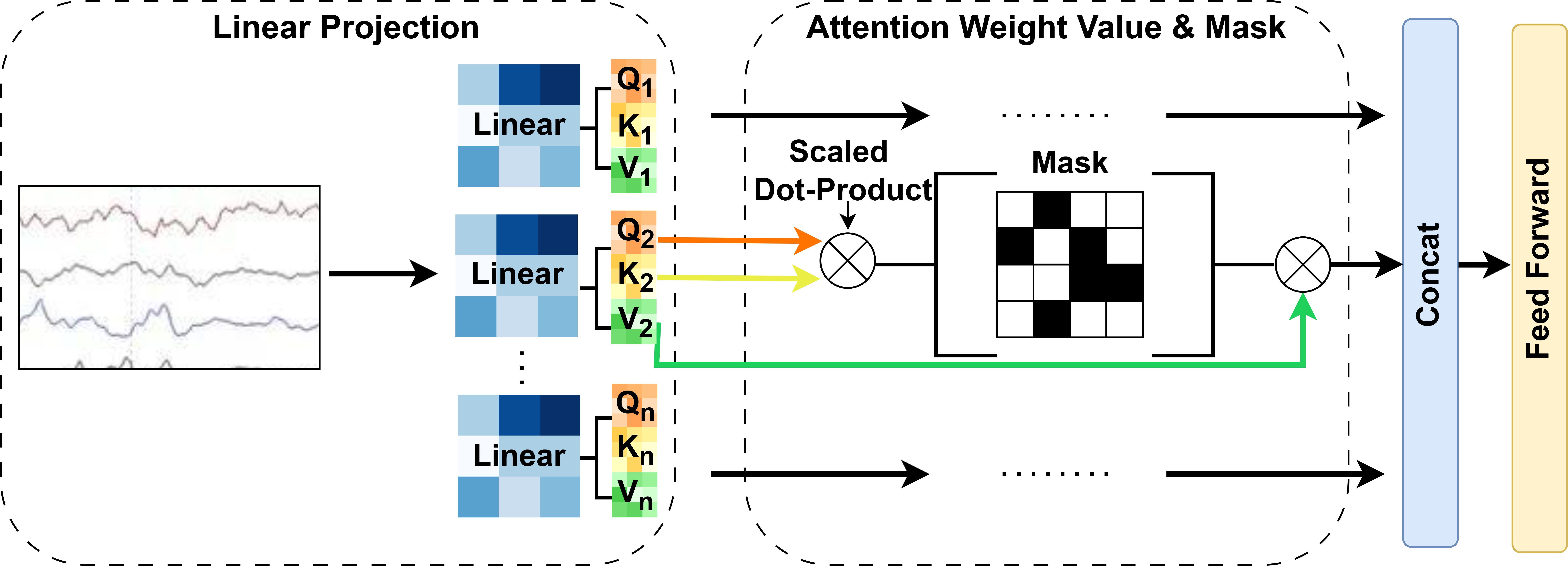} % 替换为实际图片路径
	\caption{Illustration of masking multi-head self-attention.}
\end{figure*}

\subsection{Weighted Multi-channel Filter}
Different representations of EEG signals contain not only valuable SSVEP information but also various noise components that can impede decoding. As a result of the different electrode placements across the scalp in EEG acquisition systems, the noise components exhibit variations across different channels. Therefore, by a weighted combination of channels, the components of the signal that are advantageous for SSVEP classification can be amplified, while the remaining noise components can be suppressed through mutual cancellation. Distinct from traditional spatial filters, we employed a convolution-based filtering method. N convolutional kernels of size C×1 are used in the convolutional layer to get multiple weighted results, where C represents the number of channels and N denotes the number of kernels, which is typically a multiple of C. This process can be defined as:
\begin{equation}
	Y_{i}=W_{i}^\mathrm{T}X, W_{i}\in R^\mathrm{C\times1}, X\in R^\mathrm{C\times T}
\end{equation}
where \( i \) represents the index of the convolutional kernel, \( W_i \) is a learnable parameter matrix, \( X \) is the the native or the symmetric-antisymmetric components, and \( T \) is the sequence length.

After the convolutional layer, the extracted features pass through a LayerNorm layer, an activation layer, and a dropout layer in sequence to produce the output of this module.
\subsection{Masking Multi-head Self-attention Encoder}
To enhance the modeling capability of the network, we employed a multi-head self-attention encoder\cite{b27}. Additionally, to further improve the model's performance while reducing computational complexity, we incorporated a masking mechanism. Positional encoding was first added to the output of the weighted multi-channel filtering module to maintain the sequence order information.

The encoder consists of a masking multi-head self-attention layer and a feed forward layer. In the masking multi-head self-attention layer, as described in Fig.2, the input sequence X$\in R^\mathrm{N\times T}$ are mapped to three different sequential vectors(the query $Q$, the key $K$ and the value $V$), which can be described as:
\begin{equation}
	Q=X^\mathrm{T}W_{Q},K=X^\mathrm{T}W_{K},V=X^\mathrm{T}W_{V}
\end{equation}
where $W_{Q}, W_{K}\in R^\mathrm{N\times d\_k}, W_{V}\in R^\mathrm{N\times d\_v}$ are three learnable linear martices, $d\_k$ and $d\_v$ represent dimension of query($Q$) and key($K$), dimension of value($V$) respectively. The multi-head attention mechanism linearly maps the features into multiple feature subspaces with the dimension $d\_k=N\mid n$, where n is the number of the parallel attention heads. 

In the attention layer, the attention weights are obtained by taking the dot product of $Q$ with the corresponding 
$K$. For the attention weights, we implemented a row-wise mean-based masking mechanism. Specifically, for each row in the attention weights, we calculated the mean value and used it as a threshold. Elements in the row that are smaller than the threshold were considered to have weaker relevance and are penalized by reducing their weights through masking. This process can be described as:
\begin{equation}
	\text{MaskAtten}(Q, K, V) = \text{Softmax}\left(\frac{QK^T}{\sqrt{d_k}} + \text{Mask}\right)V
\end{equation}
where $d_{k}$ is the dimension of the key vectors, used to scale the dot product, and Mask is the masking matrix obtained using the method described above, which is applied to adjust the weights at specific positions during the attention calculation. In this way, noise and other irrelevant information are filtered out, enhancing the focus of the attention mechanism while also improving computational efficiency.

\subsection{Fusion and Classifier}
For the outputs generated by the dual-perspective encoder, we fuse them at the feature level by concatenating in a token-wise manner to serve as input to the classifier. In multi-center signal processing tasks, the network may exhibit insufficient learning on certain specific signal representations. Providing multiple signal representations serves as an ensemble learning strategy, enabling the network to integrate information from various aspects, thereby improving prediction accuracy and robustness. Furthermore, due to the varying sequence lengths of different EEG components caused by factors such as sampling rate, time window, start and stop frequencies, and frequency resolution settings, this simple fusion method reduces potential information loss during data alignment, thereby enhancing the network's flexibility and stability.

The fused features are ultimately fed into a classifier based on a multi-layer perceptron (MLP) consisting of two fully connected layers. Additionally, to enhance model performance, we incorporated a LayerNorm layer, a GELU activation layer, and a Dropout layer between them.
 \begin{table}[t]
	\caption{Average Accuracy (\%) on Dataset 1}
	\begin{center}
		\begin{tabular}{c l l l} % 去掉了竖线 |
			\hline
			\multirow{2}{*}{\textbf{Method}} & \multicolumn{3}{c}{\textbf{Time Window(s)}} \\ % 去掉了竖线 |
			\cline{2-4} 
			& \textbf{\textit{0.75}} & \textbf{\textit{1.00}} & \textbf{\textit{1.25}} \\
			\hline
			FBCCA & 39.89$\pm$15.40\textsuperscript{***} & 59.39$\pm$18.51\textsuperscript{***} & 69.67$\pm$23.42\textsuperscript{***}  \\
			EEGNet & 71.55$\pm$20.70\textsuperscript{**} & 79.72$\pm$18.51\textsuperscript{**} & 86.61$\pm$14.78\textsuperscript{*}  \\
			CCNN & 71.01$\pm$20.85\textsuperscript{**} & 80.55$\pm$18.93\textsuperscript{**} & 88.05$\pm$13.77\textsuperscript{*}  \\
			SSVEPNet & 70.16$\pm$23.53\textsuperscript{**} & 80.33$\pm$19.66\textsuperscript{**} & 85.68$\pm$15.73\textsuperscript{*}  \\
			SSVEPformer & 75.06$\pm$22.20\textsuperscript{*} & 84.45$\pm$17.25\textsuperscript{*} & 87.89$\pm$14.08\textsuperscript{*}  \\
			Ours & \pmb{78.66$\pm$20.75\textsuperscript{}} & \textbf{87.81$\pm$14.22}\textsuperscript{} & \textbf{92.13$\pm$10.64}\textsuperscript{}  \\
			\hline
		\end{tabular}
		\label{tab1}
	\end{center}
\end{table}
\begin{table}[t]
	\caption{Average Accuracy (\%) on Dataset 2}
	\begin{center}
		\begin{tabular}{c l l l} % 去掉了竖线 |
			\hline
			\multirow{2}{*}{\textbf{Method}} & \multicolumn{3}{c}{\textbf{Time Window(s)}} \\ % 去掉了竖线 |
			\cline{2-4} 
			& \textbf{\textit{0.3}} & \textbf{\textit{0.4}} & \textbf{\textit{0.5}} \\
			\hline
			FBCCA & 22.56$\pm$2.55\textsuperscript{***} & 22.07$\pm$3.89\textsuperscript{***} & 23.82$\pm$4.61\textsuperscript{***}  \\
			EEGNet & 47.75$\pm$16.80\textsuperscript{**} & 51.62$\pm$19.82\textsuperscript{**} & 55.54$\pm$20.93\textsuperscript{**}  \\
			CCNN & 43.37$\pm$13.62\textsuperscript{***} & 48.75$\pm$17.57\textsuperscript{***} & 52.96$\pm$18.66\textsuperscript{***}  \\
			SSVEPformer & 49.31$\pm$17.09\textsuperscript{***} & 52.63$\pm$18.73\textsuperscript{***} & 57.09$\pm$19.58\textsuperscript{**}  \\
			Ours & \pmb{53.38$\pm$18.01\textsuperscript{}} & \textbf{57.30$\pm$20.81}\textsuperscript{} & \textbf{60.24$\pm$21.39}\textsuperscript{}  \\
			\hline
		\end{tabular}
		\label{tab2}
	\end{center}
\end{table}
\section{Experiments}
\subsection{Datasets and Preprocessing}
Two public datasets were adopted to evaluate our method.

\pmb{Dataset 1} \cite{b28}: This dataset contains EEG recordings from 10 subjects performing a 12-target SSVEP task. Stimulus frequencies ranged from 9.25Hz to 14.75Hz, with a 0.5$\pi$ phase difference between adjacent targets. EEG signals were recorded using the BioSemi ActiveTwo system with 8 occipital electrodes at a 2048Hz sampling rate. Each subject completed 15 blocks of 12 trials, with each 4-second trial downsampled to 256Hz.

\pmb{Dataset 2}: The MAMEM-SSVEP-II dataset consists of EEG recordings from 11 participants instructed to gaze at five visual stimuli flickering at frequencies of 6.66, 7.50, 8.57, 10.00, and 12.00Hz. Data were recorded using the GES 300 system with 256 electrodes at a 250Hz sampling rate and segmented into 500 trials. Following previous studies \cite{b29}, the data were band-pass filtered between 1-50Hz, and eight occipital channels (PO7, PO3, PO, PO4, PO8, O1, Oz, O2) were selected.
\subsection{Experiments Setup}
 To evaluate the zero-shot capability of our method, the leave-one-subject-out cross-validation method was adopted in our experiment. And to assess the performance of our method under different time window lengths, we conducted multiple experiments with varying time windows. The FBCCA\cite{b8}, EEGNet, CCNN, SSVEPNet, SSVEPformer were selected as baseline methods. Cross-entropy was used to compute the loss, and the Adam algorithm was employed to update the parameters, with a learning rate of 0.001 and a dropout rate of 0.5. The classification accuracy rate and information transfer rate \cite{b30} were selected as the evaluation metric in our experiment.
 \subsection{Experimental Result}
 
Table \uppercase\expandafter{\romannumeral1} presents the average accuracy of six methods on Dataset 1, while the results on Dataset 2 are shown in Table \uppercase\expandafter{\romannumeral2} (only five methods were applied to Dataset 2 since SSVEPNet is not applicable for the five-class problem). The average ITR for both datasets is presented in Table \uppercase\expandafter{\romannumeral3}. We further conducted the paired t-test between our method and other baseline methods at all time window lengths (*p$<$0.05, **p$<$0.01, ***p$<$0.001). The results indicate that our method outperforms the baseline methods in terms of both average accuracy and ITR across all time windows on both datasets, showing a significant improvement(all: p$<$0.05).

  \begin{table}[t]
 	\caption{Average ITR (bits/min) on Two Datasets}
 	\begin{center}
 		\resizebox{\linewidth}{!}{
 			\begin{tabular}{c l l l l l l} % 去掉了竖线 |
 				\hline
 				\multirow{2}{*}{\textbf{Method}} & \multicolumn{3}{c}{\textbf{Dataset 1(s)}}& \multicolumn{3}{c}{\textbf{Dataset 2(s)}}  \\ % 去掉了竖线 |
 				\cline{2-7} 
 				& \textbf{\textit{0.75}} & \textbf{\textit{1.00}} & \textbf{\textit{1.25}}& \textbf{\textit{0.3}} & \textbf{\textit{0.4}} & \textbf{\textit{0.5}} \\
 				\hline
 				FBCCA & 48.95\textsuperscript{***} & 81.60\textsuperscript{***} & 87.80\textsuperscript{***} & 0.84\textsuperscript{**} & 1.22\textsuperscript{**} & 1.76\textsuperscript{**}  \\
 				EEGNet & 151.30\textsuperscript{**} & 138.41\textsuperscript{***} & 128.99\textsuperscript{**}& 54.40\textsuperscript{*} & 71.05\textsuperscript{**} & 68.96\textsuperscript{*}  \\
 				CCNN & 149.44\textsuperscript{**} & 141.43\textsuperscript{***} & 128.03\textsuperscript{**}& 38.67\textsuperscript{**} & 58.45\textsuperscript{**} & 58.58\textsuperscript{*} \\
 				SSVEPNet & 146.12\textsuperscript{**} & 140.58\textsuperscript{***} & 126.29\textsuperscript{**}& -\textsuperscript{} & -\textsuperscript{} & -\textsuperscript{}  \\
 				SSVEPformer & 167.54\textsuperscript{*} & 154.68\textsuperscript{**} & 132.59\textsuperscript{**}& 59.33\textsuperscript{**} & 72.33\textsuperscript{**} & 71.41\textsuperscript{**}  \\
 				Ours & \pmb{182.15\textsuperscript{}} & \pmb{165.55\textsuperscript{}} & \pmb{144.04\textsuperscript{}}& \pmb{73.59\textsuperscript{}} & \pmb{92.60\textsuperscript{}} & \pmb{84.09\textsuperscript{} } \\
 				\hline
 			\end{tabular}
 		}
 		\label{tab3}
 	\end{center}
 \end{table}
 \begin{table}[t]
 	\caption{Ablation Study}
 	\begin{center}
 		\scriptsize
 		\resizebox{\linewidth}{!}{
 		\begin{tabular}{ccccc|c}
 			\hline
 			\textbf{SA stream} & \textbf{NA stream}&\textbf{WMF} & \textbf{PE} & \textbf{Mask} &\textbf{Acc}\\
 			\cline{1-6} 
 			& \checkmark  &\checkmark & \checkmark& \checkmark &83.75$\pm$18.30\\
 			\checkmark  &   & \checkmark & \checkmark &\checkmark & 85.31$\pm$17.09\\
 			\checkmark& \checkmark &  & \checkmark &\checkmark & 82.89$\pm$18.30\\
 			\checkmark  & \checkmark  & \checkmark  & &\checkmark&  86.41$\pm$16.81 \\
 			\checkmark  & \checkmark  & \checkmark  & \checkmark&&  86.02$\pm$17.53 \\
 			\checkmark  & \checkmark  & \checkmark  &\checkmark & \checkmark& \pmb{87.81$\pm$14.22} \\
 			\hline
 			%copy& More table copy$^{\mathrm{a}}$& &  \\
 			%\multicolumn{4}{l}{$^{\mathrm{a}}$Sample of a Table footnote.}
 		\end{tabular}
 	}
 		\label{ablation}
 	\end{center}
 \end{table}

 \subsection{Ablation Study}
 
 To evaluate the effectiveness of each view and module of the proposed method, we performed the ablation experiments. We sequentially removed the symmetric-antisymmetric stream (SA stream), native stream (NA stream) and each of the modules and conducted experiments on Dataset 1 with a time window set to 1 second. The results are listed in Table \uppercase\expandafter{\romannumeral4}. Ablation study reveals that both native and symmetric-antisymmetric components are useful for the classification and the modules above can improve the classification performance.

 We further validated the choice of fusion strategy on the same dataset setup. The accuracy achieved by fusing temporal and spectral features at the decision and model levels were 84.61±18.31$\%$ and 84.29±17.42$\%$, respectively, with feature-level fusion outperforming both.
\section{Conclusion}

This study proposed a compact bifocal masking attention-based method (SSVEP-BiMA) for robost SSVEP decoding which concurrently processes and integrates distinct signal representations of EEG data to get a comprehensive and generalized understanding. We validated our approach on two public datasets, and the experimental results demonstrate that our model achieves significant improvements in both accuracy and information transfer rate compared to previous methods, potentially promoting the application of BCI systems.

\end{document}